\documentclass[letterpaper, 10 pt, conference]{ieeeconf}  

\IEEEoverridecommandlockouts                              

\usepackage{cite}
\usepackage{subcaption}
\usepackage{comment}
\usepackage{amsmath,amssymb,amsfonts}
\usepackage{multirow}
\usepackage{algorithmic}
\usepackage{graphicx}
\usepackage{textcomp}
\usepackage{xcolor}
\usepackage{verbatim}
\usepackage{siunitx}
\usepackage{booktabs}
\usepackage{censor}

\newcommand{\BW}{$\pi_{\mathrm{b}}^{\mathrm{w}}$}
\newcommand{\VW}{$\pi_{\mathrm{vr}}^{\mathrm{w}}$}
\newcommand{\BP}{$\pi_{\mathrm{b}}^{\mathrm{p}}$}
\newcommand{\VP}{$\pi_{\mathrm{vr}}^{\mathrm{p}}$}

\newcommand{\VR}{VR data}

\newcommand{\methodName}{Wrench-ACT}

\newlength{\figH}
\def\BibTeX{{\rm B\kern-.05em{\sc i\kern-.025em b}\kern-.08em
    T\kern-.1667em\lower.7ex\hbox{E}\kern-.125emX}}

\title{\LARGE \bf
\methodName: Enhancing Robot Policies for Contact Rich Behavior Using Direct Wrench Control
}

\author{Johannes Hechtl*$^{1,2}$, Yannik Blei*$^{2}$, Simon Ball$^{1}$, Reihaneh Mirjalili$^{2}$, Michael Krawez$^{2}$,\\ Seongjin Bien$^{2}$, Philipp Schmitt$^{1}$, Wolfram Burgard$^{2}$
\thanks{*Equal Contribution}
\thanks{$^{1}$Siemens Research and Predevelopment}%
\thanks{$^{2}$Department of Computer Science and Artificial Intelligence, University of Technology Nuremberg, Germany}%
}

\begin{document}

\maketitle

\thispagestyle{empty}
\pagestyle{empty}

\begin{abstract}

While contact-rich manipulation requires deliberate regulation of interaction forces, recent approaches to robot manipulation learning predominantly represent actions as target positions or poses. Even methods that incorporate force sensing either use it solely as an observation or, when predicting forces as part of the output, rely on a hybrid force controller.
In this paper, we propose an imitation learning policy that predicts wrenches as its sole action output for direct use by a pure force controller.
Our studies suggest that force-domain imitation learning depends critically on data collection, with force-feedback teleoperation improving policy performance by capturing the operator's deliberate force regulation. 
Using Action Chunking with Transformers (ACT) as the base architecture, we train single-task models on bilateral wrench demonstrations and evaluate them on five contact-rich manipulation tasks. The wrench policy matches or outperforms position-based baselines across all tasks, with gains varying according to the degree of deliberate force regulation each task requires.
Cross-condition ablations show that the bilateral data collection interface and the wrench action space each contribute independently to performance. To support further research, we will release over 1{,}000 wrench-action demonstrations spanning these tasks on a companion website upon publication.

\end{abstract}

\section{Introduction}
Many manipulation tasks of industrial relevance involve deliberate regulation of contact forces. Many assembly operations, including peg insertion under tight tolerances and connector mating, require not only geometric precision but also controlled interaction forces; applying too little leaves the task incomplete, while applying too much damages parts or breaks the assembly.

Despite this, force typically appears in robot learning policies on the input side as an observation that the policy can condition on, but not as a quantity it plans. A natural alternative is to place force on the output side and have the policy directly predict the desired end-effector wrench as its action. We argue this is fundamentally more capable for contact-rich tasks. A policy that outputs wrenches can explicitly plan what forces to apply, whereas a policy that outputs positions can only produce forces as an emergent side effect of the controller.
We consider this work as a viability study of the wrench action space, examining whether it yields competitive policies for contact-rich manipulation. To this end, we train an ACT-based~\cite{zhao2023learning} wrench policy, and evaluate it on five contact-rich tasks. 

As with any imitation learning approach, the quality of the training data puts an upper limit on policy performance. For a wrench policy, we hypothesize that data should contain deliberately commanded forces. However, demonstrations collected via position-based teleoperation do not provide this. Instead, the operator commands gripper poses, and any contact forces that arise are incidental consequences of the impedance controller tracking a position command, not deliberate operator intent. Bilateral force-reflecting teleoperation addresses this directly by applying forces to a leader arm, which are mirrored by the follower, and the follower's motion is reflected back as haptic feedback. Thus the operator commands wrenches, not positions, and the recorded demonstrations contain intentional force behavior as the action signal. Our experiments confirm that training on data from teleoperation with haptic feedback improves wrench policies compared to training on forces derived from position commands.

\begin{figure}[t]
    \centering
    \includegraphics[clip,trim=0.5cm 0.5cm 0.5cm 0.5cm,width=1\linewidth]{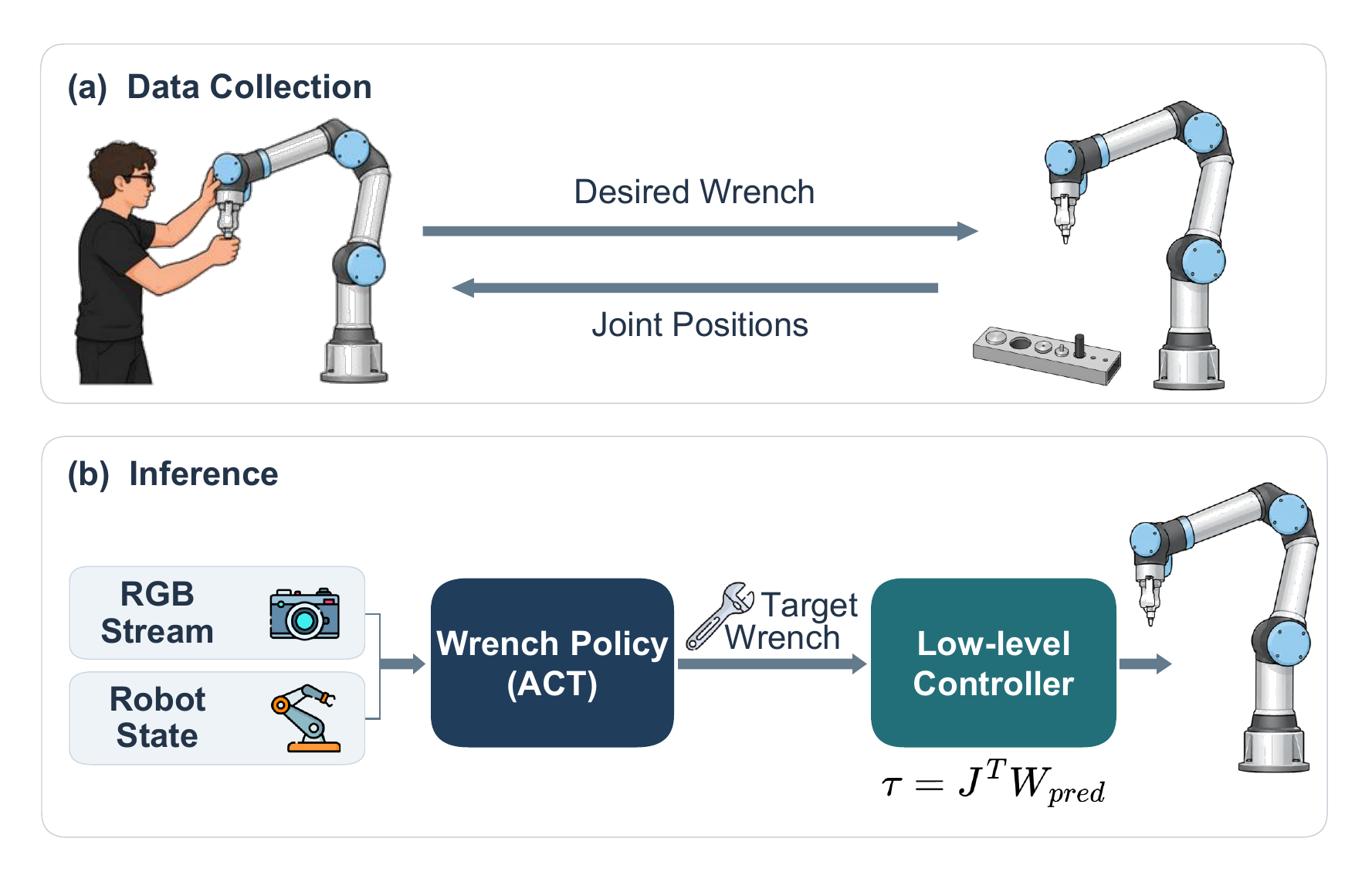} 
    \caption{Method overview of the proposed pure wrench approach. \textbf{(A)} Data collection via bilateral teleoperation. Operator wrench commands are executed by the follower arm and recorded, while the follower position is mirrored back to the leader. \textbf{(B)} Inference. Multimodal observations (RGB streams and robot state) are processed by the policy to directly predict a 6D target wrench, entirely bypassing intermediate positional control for contact-rich manipulation.}
    \label{fig:overview}
\end{figure}

\noindent\textbf{In summary, this paper makes the following contributions:}
\begin{enumerate}
  \item A novel pipeline combining bilateral force-reflecting teleoperation with a wrench-action policy, aligning the data collection interface with the action space.
  \item A systematic empirical study of the wrench action space across five contact-rich tasks, including cross-condition ablations.
  \item A dataset of over 1{,}000 wrench-action demonstrations across tasks, released to support future work.
\end{enumerate}

\section{Related Work}

Our work sits at the intersection of three research threads: how force information is incorporated into robot learning policies, how the data collection interface shapes the demonstrations used for training, and whether force or wrench can serve as the policy's output rather than its input. The compliance and impedance literature forms a natural bridge between the first and third threads, since variable-impedance methods partially close the gap between position-based and force-based action spaces, and understanding their limitations motivates the simpler, assumption-free approach we take.

\subsection{Force and Torque as Policy Input}
A large body of work improves contact-rich manipulation by incorporating force or torque on the observation side while leaving the action space positional. TA-VLA~\cite{TA-VLA} injects joint torque into a VLA decoder, finding that a single compressed torque-history token outperforms multi-token representations. ForceVLA~\cite{yu2026forcevla} fuses force tokens with vision-language tokens via a mixture-of-experts module. FoAR~\cite{he2025foar} encodes a temporal force history with a Transformer and gates its influence through a learned contact predictor. ManipForce~\cite{manipforce} pairs a handheld RGB-F/T collection device with a frequency-aware Transformer that preserves high-frequency force signals through cross-attention fusion.

A complementary line of work replaces wrist-mounted F/T sensors with tactile or acoustic signals. TAP-VLA~\cite{tap-vla} overlays tactile shear vectors onto RGB frames without modifying the pretrained VLA architecture. T-Rex~\cite{trex} combines low-frequency visuomotor planning with high-frequency tactile refinement. ManiWAV~\cite{maniwav} and Yi \emph{et al.}~\cite{soundoftouch} recover contact properties from piezoelectric microphones and vibration spectrograms, respectively. FD-VLA~\cite{zhao2026fd} and FARM~\cite{helmut2025tactile} also fall into this category.

In all of these methods, the action remains a motion command, and contact forces are emergent rather than planned.

\subsection{Compliance and Impedance Parameter Prediction}
Contact-rich tasks require the robot to be stiff when precise positioning matters and compliant when it must yield to contact forces, motivating methods that adapt impedance parameters to the current interaction state rather than fixing them at design time.
Learning variable impedance from demonstration has a long history~\cite{abudakka2020variable}. Early approaches used probabilistic motion primitives to jointly encode trajectory and time-varying stiffness from kinesthetic teaching~\cite{rozo2016learning}, later extended to compliance learned directly from force measurements~\cite{abudakka2018force}. Kronander and Billard~\cite{kronander2016stability} established stability conditions for variable impedance controllers learned from data.

More recent works embed these ideas into end-to-end policy learning. ACP~\cite{AdaptiveCompliancePolicy} jointly predicts end-effector pose and Cartesian stiffness via a diffusion policy conditioned on F/T spectrograms. FILIC~\cite{GE2025FILIC} uses a dual-loop impedance architecture with a real-time MuJoCo digital twin and URF~\cite{shin2026urf} predicts a full stiffness matrix together with an impedance/admittance switch ratio. UniForce~\cite{chen2026uniforce}, PhaForce~\cite{wang2026phaforce}, and phase-conditioned methods~\cite{chen2026phaseconditioned} combine compliance adaptation with discrete contact phases that are manually labeled per task. Xu \emph{et al.}~\cite{xu2026mindthegap} instead recover implicit impedance from the mismatch between operator intent and executed trajectory, and Shukla \emph{et al.}~\cite{shukla2026hierarchical} decouple a slow diffusion-based planner from a fast neural impedance controller.

A structural cost shared across all of these methods is reliance on task-specific parameters that cannot be derived from demonstrations alone. Stiffness bounds, contact-phase labels, mode-switch thresholds, and damping coefficients must be hand-tuned or estimated through specialized procedures for each new task. Our approach avoids this entirely by predicting the wrench directly. That requires no such parameters, and the same pipeline applies across tasks without per-task engineering.

\subsection{Teleoperation for Data Collection}
Position-based teleoperation systems have become the dominant interface for collecting robot learning demonstrations. ALOHA~\cite{zhao2023learning} popularized low-cost leader-follower setups using joint position mirroring, while GELLO~\cite{wu2024gello} extends this to a wider range of robot hardware with minimal calibration overhead. The Universal Manipulation Interface (UMI)~\cite{chi2024umi} pushes further, enabling demonstration collection in unstructured environments using handheld grippers. These systems excel at capturing geometric trajectories but share a structural limitation: the operator commands positions, so any contact forces in the recorded data are artifacts of position tracking rather than deliberate commands.

Bilateral teleoperation addresses this by coupling a leader and a follower through both motion and force channels. Here, the operator's applied forces are reflected to the follower, and the follower's contact forces are fed back haptically to the operator~\cite{hannaford1989design,lawrence1993stability}. This bidirectional coupling gives the operator direct wrench authority and produces demonstrations in which contact forces are deliberately commanded rather than incidental. Adachi \emph{et al.}~\cite{adachi2018imitation} showed that bilateral control can serve directly as a data collection interface for imitation learning, recording synchronized position and force signals for contact-rich tasks, and Sakaino~\cite{sakaino2021bilateral} extended this to velocity-controlled robots by converting predicted force references into velocity commands via admittance control. On the handheld-device side, TacUMI~\cite{tacumi2026} extends the UMI gripper with a dedicated force/torque sensor and tactile sensing, enabling synchronized force and vision recording without a robot arm during collection. Our setup follows the bilateral paradigm. The operator commands wrenches through the leader arm, and follower position is reflected back haptically, yielding demonstrations with intentional force behavior as the action signal.

\subsection{Direct Force and Wrench Prediction}
\label{sec:directforce}
Closest to our work, a handful of methods predict force or wrench targets directly, thus going beyond compliance parameter prediction to specify explicit force goals. ForcePolicy~\cite{fang2026forcepolicy} uses a two-level architecture where a high-frequency local policy estimates the interaction frame and executes hybrid force-position control~\cite{raibert1981hybrid}, with contact modes manually labeled per task. TactileVLA~\cite{huang2025tactilevla} activates force-control knowledge latent in a VLA through a hybrid controller, requiring a manually set force dead-band and gain matrix. ForceMimic~\cite{liu2025forcemimic} jointly predicts end-effector position and contact force, where the predicted force gates a hard switch between a pure IK controller and a hybrid force controller at a fixed 6\,N threshold tuned per task. ForceVLA2~\cite{li2026forcevla2} outputs a 7D pose delta and a 6D force/torque command simultaneously, with the force shaping execution via a Jacobian-based mapping but the pose delta remaining the primary action. Bi-ACT~\cite{biact2024} extends the ACT architecture with bilateral teleoperation, predicting joint-space forces alongside joint positions and feeding them back to the operator as haptic signals.

In all of these methods, force is a secondary, it modulates or gates a positional action rather than being the commanded quantity itself. In contrast, we predict the full 6D task-space wrench as the \emph{sole} action, with no positional output and no per-task force parameters.

\begin{figure*}[t]
    \centering
  \begin{subfigure}[t]{0.3\textwidth}
    \includegraphics[width=\linewidth]{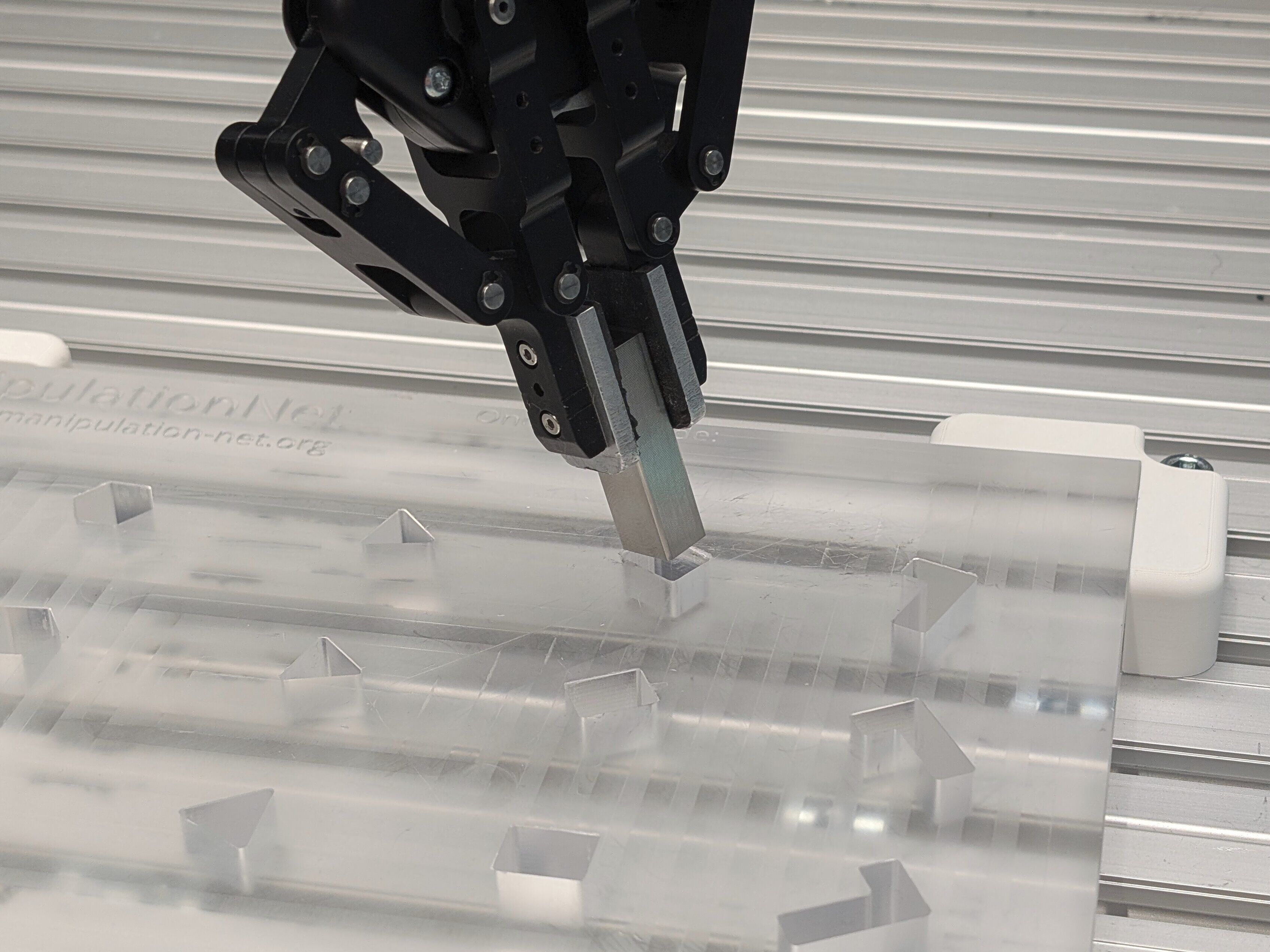}
    \caption{Peg Insertion}
  \end{subfigure}\hfill
  \begin{subfigure}[t]{0.3\textwidth}
    \includegraphics[width=\linewidth]{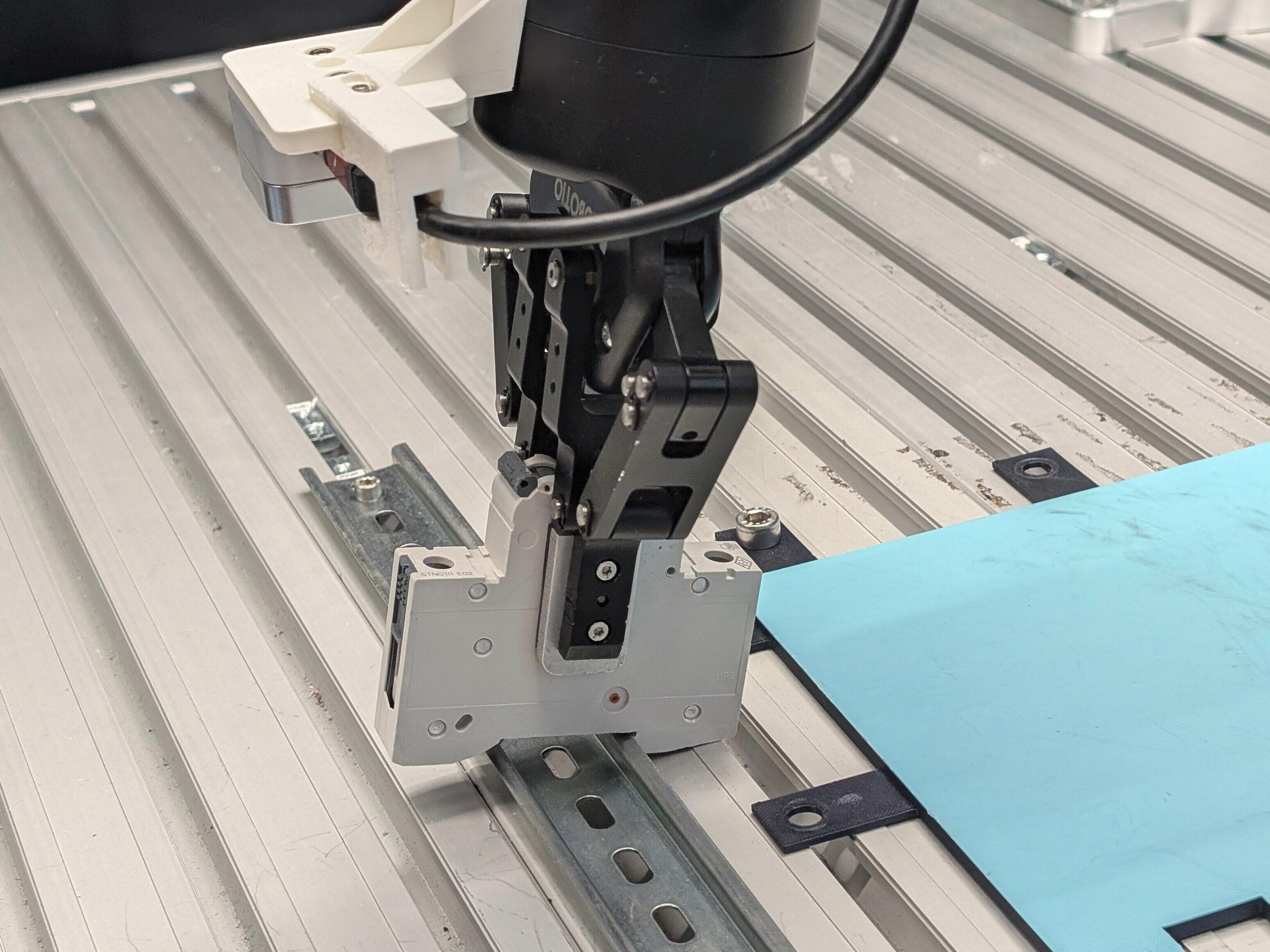}
    \caption{Fuse Clipping}
  \end{subfigure}\hfill
  \begin{subfigure}[t]{0.3\textwidth}
    \includegraphics[width=\linewidth]{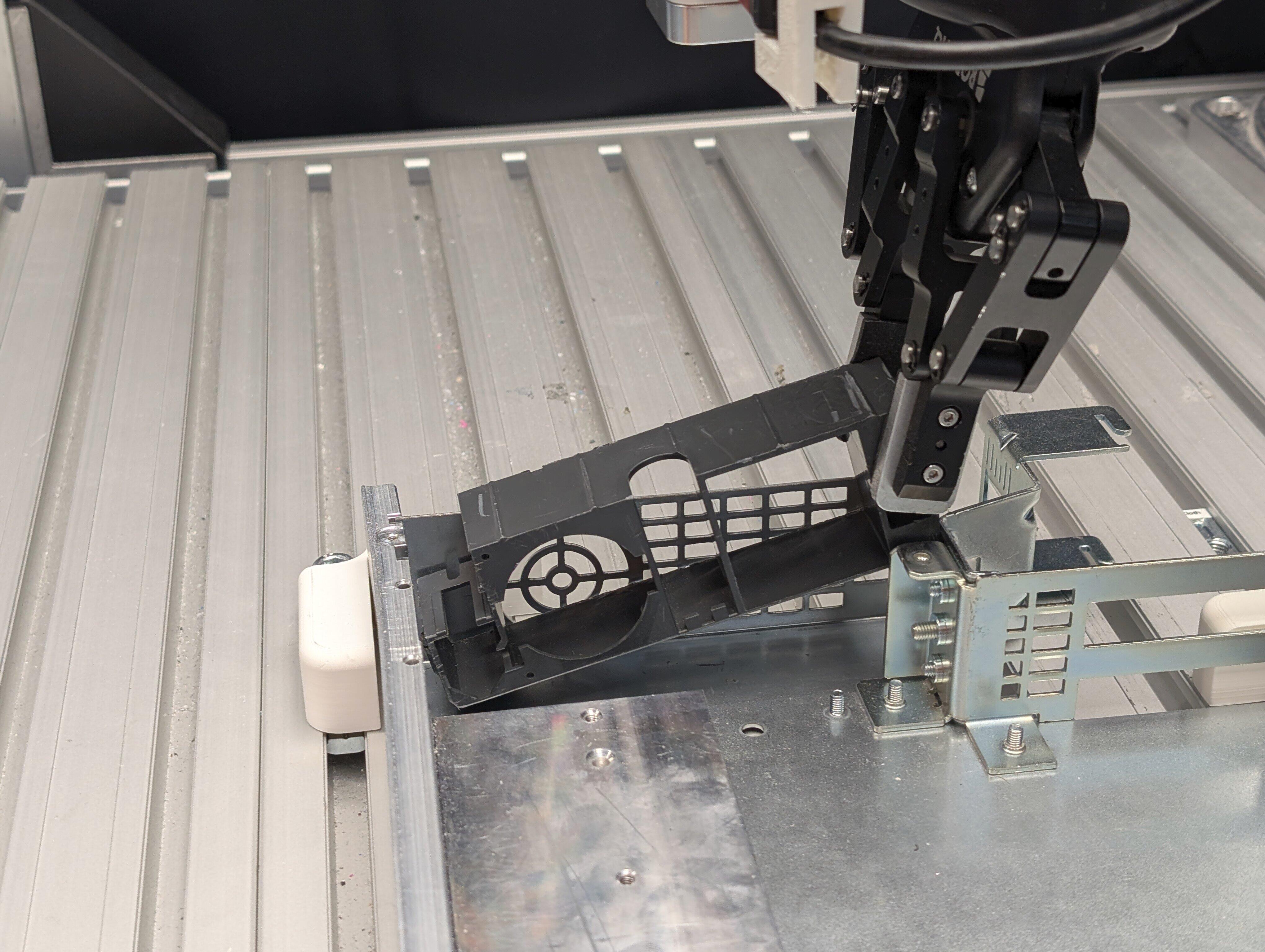}
    \caption{Fan Insertion}
  \end{subfigure}\hfill
  \par\vspace{0.5em}
  \begin{subfigure}[t]{0.3\textwidth}
    \includegraphics[width=\linewidth]{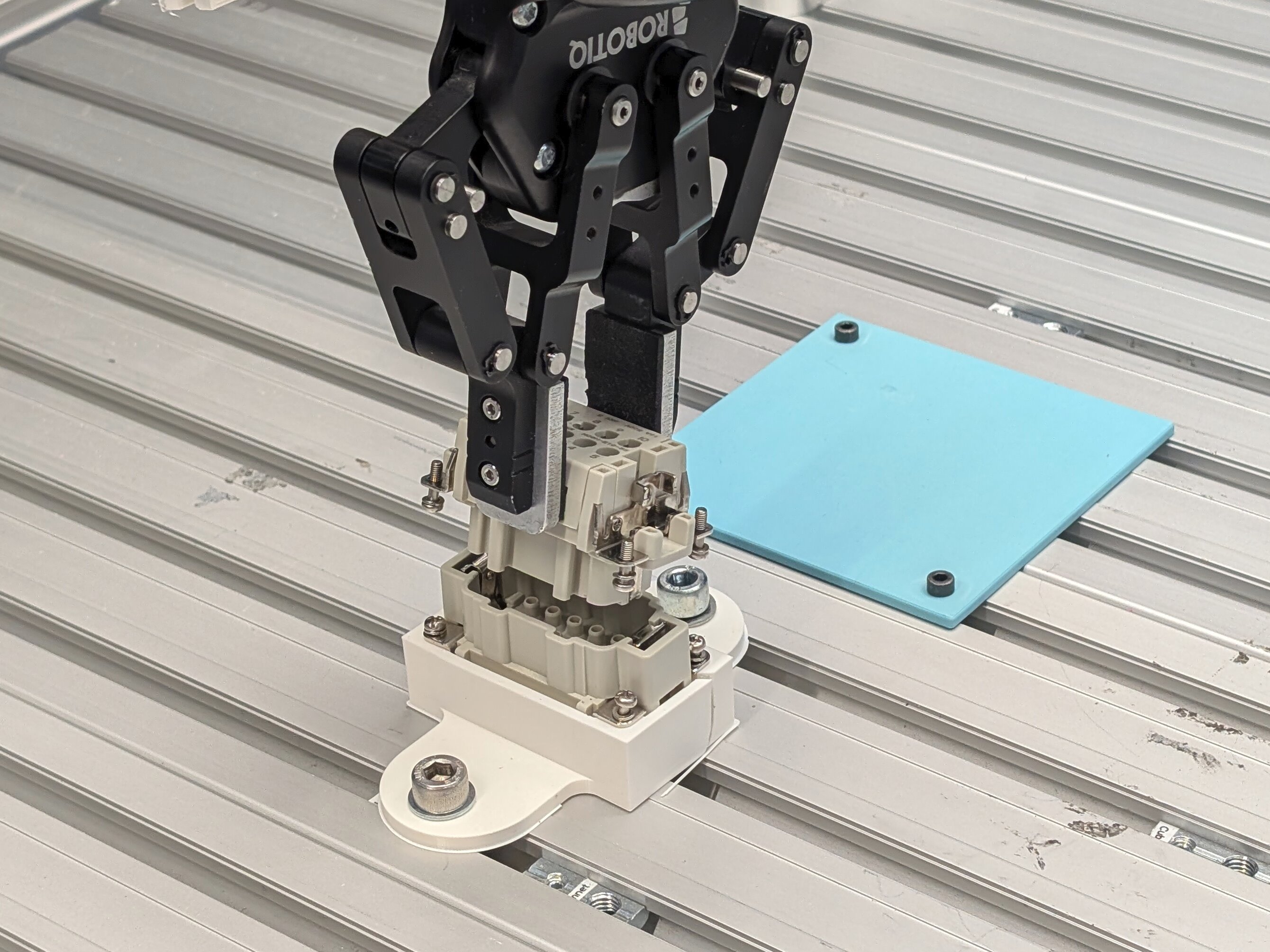}
    \caption{Industrial Connector}
  \end{subfigure}\hfill
  \begin{subfigure}[t]{0.3\textwidth}
    \includegraphics[width=\linewidth]{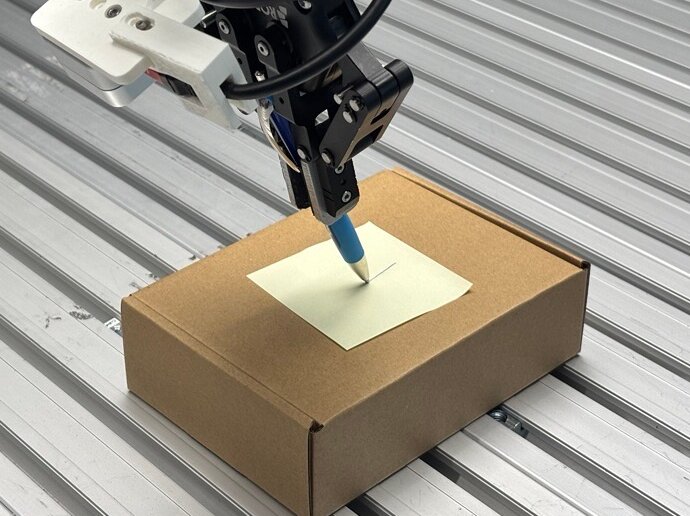}
    \caption{Pen Writing}
  \end{subfigure}\hfill
  \begin{subfigure}[t]{0.3\textwidth}
    \includegraphics[width=\linewidth]{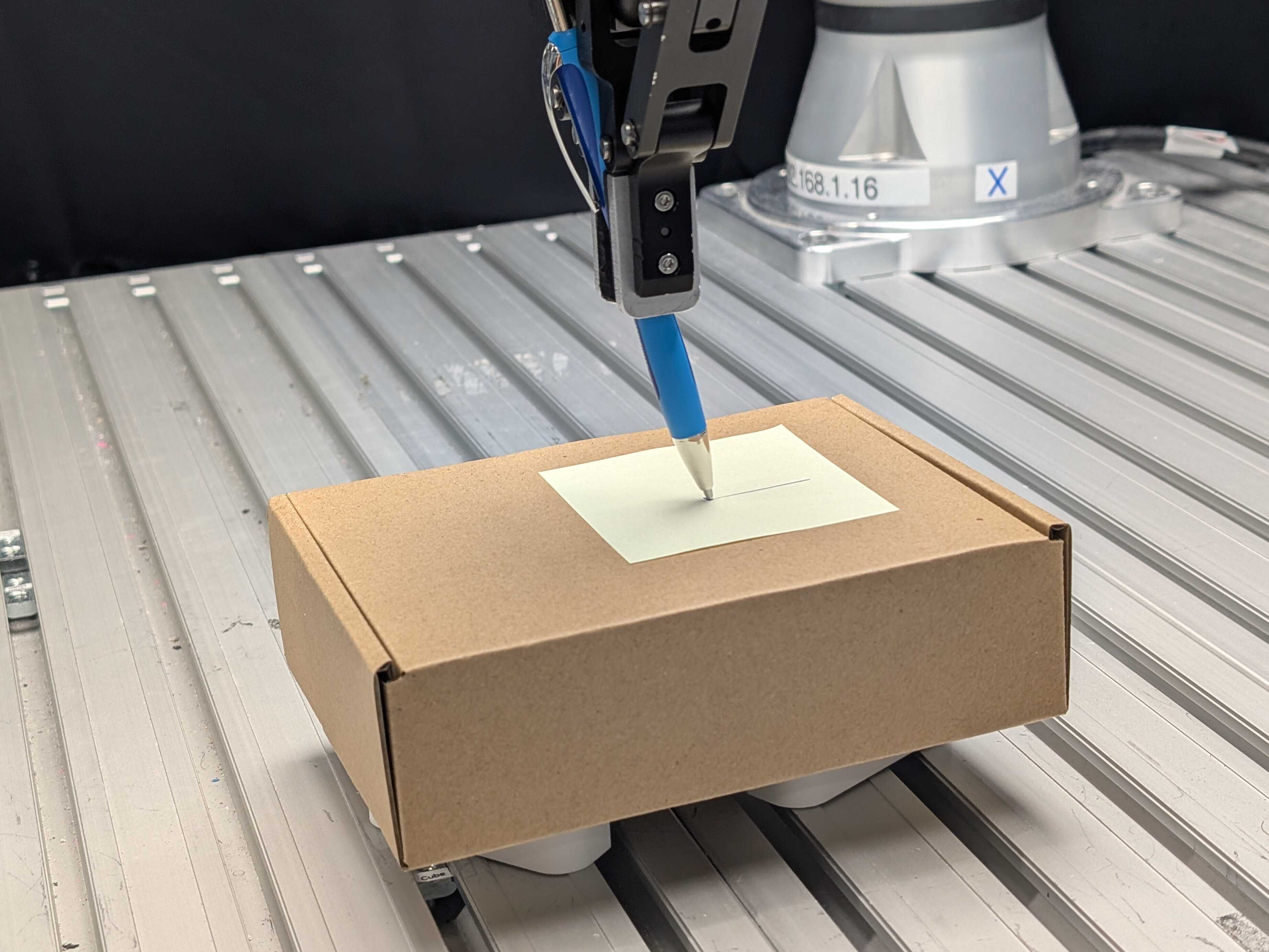}
    \caption{Ablation \ref{sec:ablation_box_height}: Raised Pen Writing}
  \end{subfigure}
  \caption{Overview of the five contact-rich manipulation tasks evaluated in this work (a - e) and the ablation evaluating robustness to box-height variation (f)}
    \label{fig:tasks}
\end{figure*}

\section{Method}

\subsection{Direct Wrench Control}

To clarify how our action space differs from position-based approaches, we begin with a standard Cartesian impedance controller:

\begin{equation}
\begin{aligned}
\mathbf{w}_{\text{cmd}}
&=
K_p(x_{\text{des}} - x)
+
K_d(\dot{x}_{\text{des}} - \dot{x})
+ \mathbf{w}_{\text{des}}, \\
\tau
&=
J(q)^\top \mathbf{w}_{\text{cmd}}
+
N(q)^\top \tau_{\text{null}}
\end{aligned}
\label{eq:impedance}
\end{equation}
where $x$ and $\dot{x}$ denote the current end-effector pose and velocity, while $x_{\text{des}}$ and $\dot{x}_{\text{des}}$ are the corresponding desired values. The stiffness $K_p$ converts pose error into a restoring wrench, and the damping $K_d$ opposes relative motion. An optional feedforward wrench $\mathbf{w}_{\text{des}}$ can be applied independently of pose error. The resulting commanded wrench $\mathbf{w}_{\text{cmd}}$ is mapped to joint torques $\tau$ through the transpose of the robot Jacobian $J(q)$; the null-space term controls redundant motion without changing the end-effector command.

Most learned manipulation policies act through the first term of~\eqref{eq:impedance} by predicting $x_{\text{des}}$ and contact forces arising only when the controller encounters an obstacle and develops a pose error. Adaptive-compliance methods such as ACP~\cite{AdaptiveCompliancePolicy} additionally predict $K_p$, allowing the policy to regulate how strongly pose errors are converted into force. Methods that also predict a wrench, such as ForceVLA2~\cite{li2026forcevla2}, retain a positional action and use the wrench output to shape or arbitrate its execution. They therefore still require both motion and force-control pathways.

Prior work has shown that reducing stiffness can improve learned contact-rich manipulation~\cite{AdaptiveCompliancePolicy,bronars2026tune}, because a compliant robot can yield to geometric uncertainty instead of forcing the commanded pose. We take this trend to its limiting case by setting $K_p=0$ and omitting the desired pose entirely. Our policy directly predicts $\mathbf{w}_{\text{des}}$, while the damping term remains as a fixed stabilizing component. Consequently, the policy specifies the wrench to apply rather than a pose error from which the controller must indirectly generate one. The same action representation is used in free space and in contact, without switching between controllers.

\subsection{Policy Architecture}

We adopt ACT~\cite{zhao2023learning} as our base architecture without architectural modification, and we refer the reader to the original paper for details. Since no large wrench-action datasets exist, we train single-task models from scratch on our collected demonstrations. ACT is well-suited to this regime, having proven effective for contact-rich manipulation from limited data. Its low inference latency allows the policy to run at approximately \SI{50}{\hertz}. Force control is widely assumed to require high-frequency policy updates. We therefore examine this assumption empirically in Section~\ref{sec:ablations}.

The observation space consists of three synchronized RGB camera streams together with the robot's Cartesian end-effector position, Cartesian end-effector velocity, the 6D wrench measured at the follower's end-effector, and a scalar gripper position. The action space is a seven-dimensional vector comprising the 6D target force-torque and a scalar gripper opening command.


\subsection{Data Collection}
\label{sec:dataCollection}
We use a bilateral teleoperation setup with a leader and a follower robot arm. Both arms are equipped with F/T sensors. The operator grasps the leader arm and applies force to it. This force is then measured as a 6D wrench and mirrored by the follower arm. The position of the follower arm is then mirrored back to the leader.
Thus, the operator controls the \textit{wrench}, not the position, and receives force feedback through the mirrored position of the follower.
We record the commanded wrench (F/T measurement at the leader), the measured wrench at the follower's end effector, the follower position, and the follower velocity.
We use the UR5e robot, commanding the desired wrench via the robot's built-in force mode.

\subsection{Inference}
At execution time, the predicted target wrench is sent directly to the UR5e's force mode, which applies the commanded 6D wrench at the end-effector. Cameras, robot kinematics, and force measurements are all acquired at \SI{60}{\hertz}, while the underlying robot and force control loops run at \SI{500}{\hertz}. Policy inference runs at approximately \SI{50}{\hertz} with the most recently predicted wrench being held constant between inference steps.

Because the gripper assembly has a mass of approximately \SI{1}{\kilogram}, a small commanded force produces only a small free-space acceleration, which inherently limits unintended motion between policy steps. To further improve stability, we add a velocity-proportional damping term to the commanded wrench and obtain the final control law
\begin{equation}
    \mathbf{w}_{\text{cmd}}=\mathbf{w}_{\text{pred}}-K_d \dot{x}.
\end{equation}

\section{Experimental Evaluation}

\subsection{Tasks and Setup}
All tasks are purposefully contact-rich. For manipulation tasks where contact forces are incidental, e.g., pick-and-place, we do not expect wrench control to offer meaningful advantages over position control, since the policy has no force behavior to learn. Our evaluation therefore focuses on tasks where we expect explicit force regulation to matter for success.

We evaluate on five contact-rich manipulation tasks (see Figure~\ref{fig:tasks}):
\textbf{(1)~Peg Insertion:} a rectangular peg must be inserted into a hole with \SI{0.02}{\milli\meter} tolerance, as specified by the ManipulationNet~\cite{manipulationnet} benchmark. The peg is grasped by the robot, but during rollout an operator holds the peg and ensures a good grasp, so only the transfer motion and the insertion are performed autonomously.
\textbf{(2)~Fuse Clipping:} a Siemens fuse must first be picked up from a small area and then clipped onto a top-hat rail, requiring alignment and a firm snap-in force.
\textbf{(3)~Fan Insertion:} a plastic fan part must be picked up and mounted onto a Siemens inverter's metal casing. The task is challenging because the plastic part easily jams or snags on the edges of the metal structure.
\textbf{(4)~Industrial Connector:} the robot grasps a male connector and inserts it into a fixed female counterpart, requiring precise spatial alignment and substantial insertion force.
\textbf{(5)~Pen Writing:} the robot holds a pen and draws a line on paper resting on a cardboard box. This task requires careful force regulation, since too little contact pressure and the pen does not write, too much and the pen punctures the cardboard.

\begin{figure}
    \centering
    \includegraphics[width=0.9\linewidth]{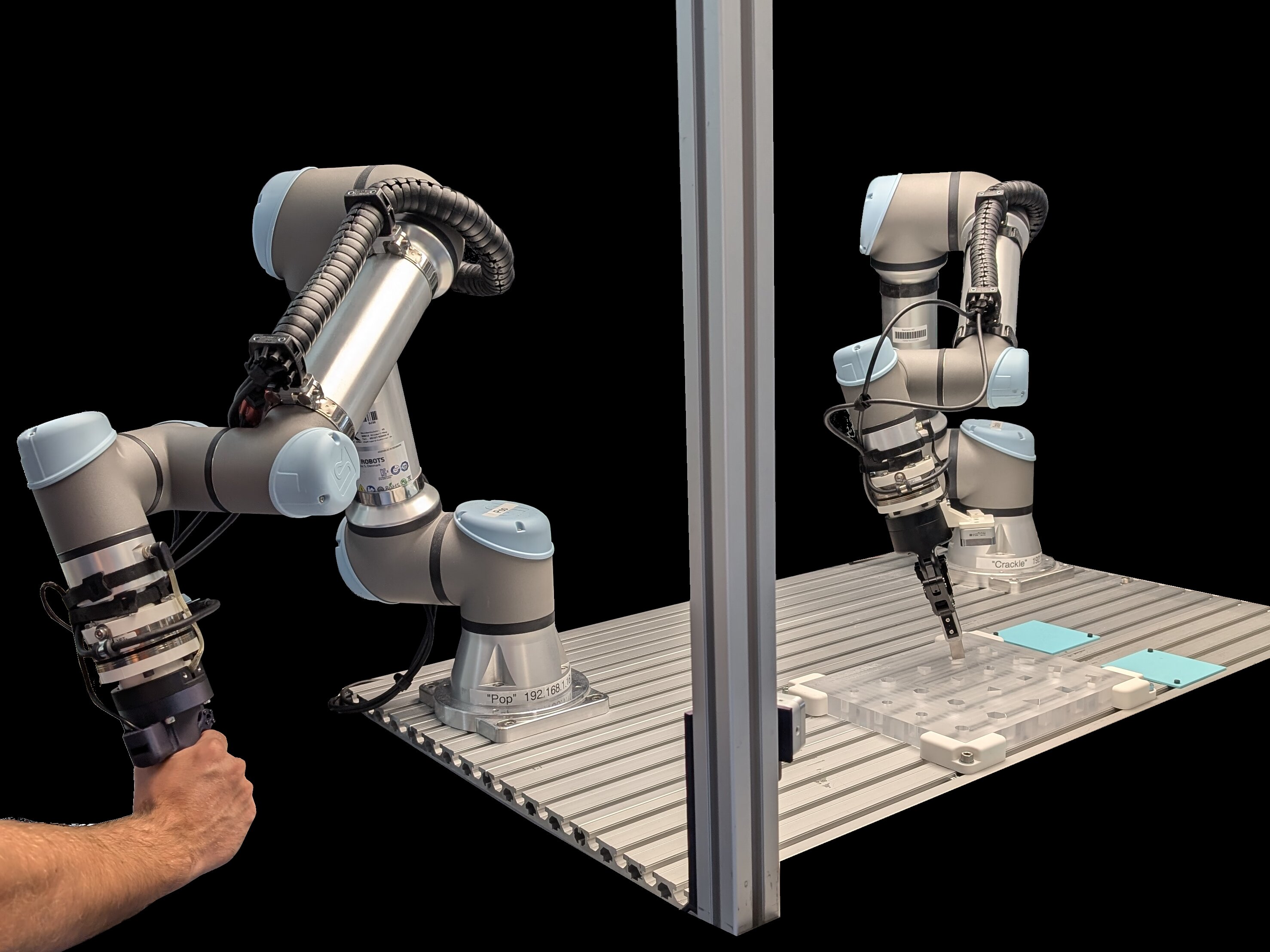}
    \caption{Bilateral Teleoperation Setup. Operator wrench commands, measured by the leader (left), are executed by the follower arm (right). The follower position is mirrored back to the leader. }
    \label{fig:wrench_teleop_setup}
\end{figure}

For our wrench action space recording, we use two Universal Robots UR5e robotic arms (see Figure~\ref{fig:wrench_teleop_setup}). One robot serves as a leader, the other as a follower. Each one is equipped with a Schunk FTN AXIA 80 6-DOF Force/Torque sensor to accurately measure target and interaction wrench respectively. The UR5e does not have torque sensors and cannot produce accurate torques, we instead use the Force Mode of the UR5e controller which uses an admittance controller and the readings of the external Force-Torque sensor to achieve the desired end-effector wrench.

When recording in a position action space, we use one UR5e arm and a Meta Quest 3 VR headset for teleoperation.

For inference, we deploy a workstation with an NVIDIA RTX5090 GPU. 
Three Intel RealSense D405 cameras provide coverage of the scene with two static cameras at the top and to the side and a wrist camera.

\begin{figure*}[t]
    \centering
    \includegraphics[width=\linewidth]{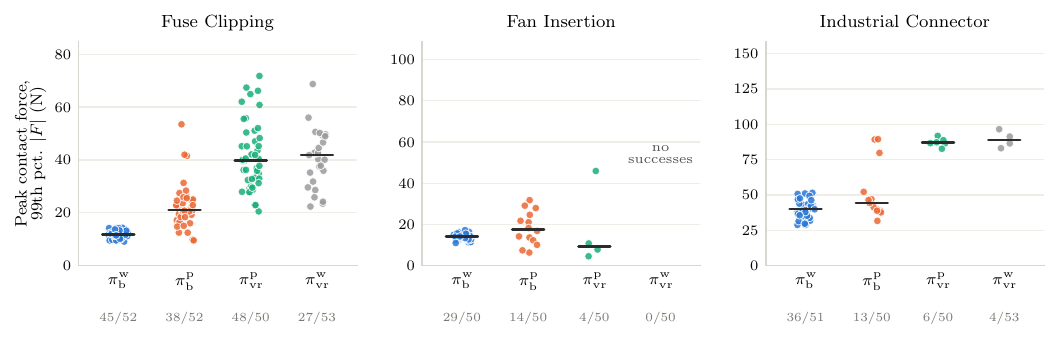}
    \caption{Results of the contact force analysis. Bilateral teleoperation with a wrench action space (\BW) achieves high success rates while maintaining relatively low median 99th-percentile force magnitude and a narrow distribution of applied contact forces.}
    \label{fig:contactForce}
\end{figure*}

\subsection{Datasets and Policies}
A key aspect of our evaluation is the decoupling of the data collection interface from the action representation. We collect two datasets per task in the following manner:

\textbf{Bilateral data} is collected via the bilateral force-reflecting teleoperation setup described in Section~\ref{sec:dataCollection}. The operator commands wrenches directly, recorded actions are the 6D wrenches measured at the leader arm.

\textbf{VR data} is collected via a Meta Quest VR controller driving the robot through a Cartesian impedance controller, the standard position-based teleoperation setup used in most prior work. Recorded actions are end-effector poses.

To disentangle the effect of the collection interface from the action representation, we additionally convert each dataset into the other action space. Wrench actions $w_{\text{cmd}}$ for the VR dataset are computed from the recorded trajectories and contact forces via the impedance model $w_{\text{cmd}} {=} K_p(\Delta_{\text{pos}}) {+} K_d {\cdot} \Delta_{\text{vel}}$. It applies a stiffness $K_p$ on the positional error $\Delta_\text{pos}=x_{des}-x$ and a damping $K_d$ on the velocity error $\Delta_\text{vel} = \dot{x}_{des} - \dot{x}$. Position actions $\Delta_{\text{pos}}^*$ for the bilateral dataset are extracted in the opposite direction by inverting the impedance model: 
\begin{equation}
    \Delta_{\text{pos}}^* = \arg\min_{\Delta_{\text{pos}}} \left\| K_p(\Delta_{\text{pos}}) + K_d \cdot \Delta_{\text{vel}} - \mathbf{w}_{\text{cmd}} \right\|_2.
\end{equation}

This yields four policies per task, summarized in Table~\ref{tab:policies}.

All tasks are contact-rich and are only achievable with a compliant controller. For the impedance controller used as baseline (\VP), we tune stiffness to be as low as possible while avoiding the large phase lags that would make teleoperation feel sluggish or unintuitive to the operator. This choice is motivated by prior work showing that lower stiffness improves success on contact-rich tasks~\cite{AdaptiveCompliancePolicy,bronars2026tune}, and it ensures the comparison is fair by configuring the impedance baseline as recommended by prior work.

\begin{table}[b]
  \centering
  \caption{The four policies resulting from the combination of data collection interface and action representation.}
  \label{tab:policies}
    \renewcommand{\arraystretch}{1.6}
  \begin{tabular}{llcc}
  \toprule
  \multicolumn{2}{c}{\textbf{Action Space}}
  & Wrench & Position \\
  \midrule
  \multirow{2}{*}{\textbf{Dataset}} & Bilateral & \BW & \BP \\
                                    & \VR        & \VW & \VP \\
  \bottomrule
  \end{tabular}
\end{table}

\noindent We collect between 200 and 300 episodes per task, making sure that for each task the same number of VR and wrench action episodes are recorded. All policies are trained using ACT~\cite{zhao2023learning} for 150{,}000 steps with a batch size of 32.

\subsection{Results}

\begin{table}[t]
  \centering
  \caption{Success rates (\%) across tasks and policies. \BW: bilateral data with wrench actions; \BP: bilateral data with pose actions; \VW: VR data with wrench actions; \VP: VR data with pose actions. All policies are evaluated on 50 rollouts for all tasks.}
  \label{tab:results}
  \begin{tabular}{lcccc}
  \toprule
  \textbf{Task} & \BW & \BP & \VW & \VP \\
  \midrule
  Peg Insertion        & \textbf{74\%} & 2\% & 2\% & 6\% \\
  Fuse Clipping        & 88\% & 72\% & 52\% & \textbf{96\%} \\
  Fan Insertion        & \textbf{58\%} & 28\% & 0\% & 8\% \\
  Industrial Connector & \textbf{80\%} & 26\% & 4\% & 12\% \\
  Pen Writing          & 84\% & 46\% & \textbf{86\%} & 78\% \\
  \midrule
  Average              & \textbf{76.8}\% & 34.8\% & 28.8\% & 40\% \\
  \bottomrule
  \end{tabular}
\end{table}

\begin{figure}[t]
    \centering
    \includegraphics[width=0.9\linewidth]{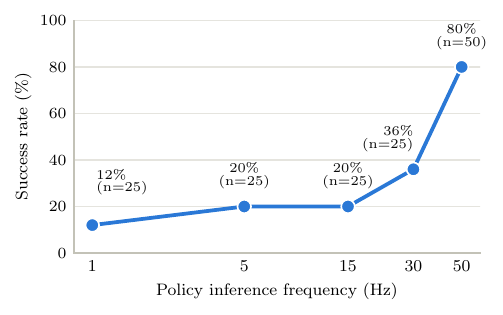}
    \caption{Ablation on the inference frequency on the industrial connector task. We show the dependence of success rate on inference speed. Contrary to a widespread assumption, an inference rate of \SI{50}{Hz} is sufficient for a success rate of 80\% in our setting.} 
    \label{fig:frequency}
\end{figure}

Results on task success are summarized in Table~\ref{tab:results}. We evaluate all policies with 50 rollouts on each task. We report 95\% confidence intervals and use two-proportion $z$-tests (Fisher's exact test where any expected cell count is below~5) with Bonferroni correction for the five tasks ($\alpha' = 0.01$).

Our approach, the bilateral wrench policy (\BW) matches or outperforms the position baseline (\VP) with average success rates over all tasks being 76.8\% and 40\%, respectively. The difference is significant on Peg Insertion, Fan Insertion, and Industrial Connector, where \BW{} achieves 74\%, 58\%, and 80\% against 6\%, 8\%, and 12\% for \VP{}. On Fuse Clipping and Pen Writing, the difference is not significant. 

Within the wrench action space, bilateral teleoperation achieves an average success rate of 76.8\% compared to 28.8\% for VR collection, significantly outperforming on four of five tasks and showing similar performance for pen writing. Within the position action space, the picture reverses: \VP{} significantly outperforms \BP{} on Fuse Clipping and Pen Writing (40\% vs.\ 34.8\% overall). Our bilateral interface is therefore not universally better as a data collection method, but it specifically benefits the wrench action space.

Within bilateral data, the wrench action space significantly outperforms position actions on four of five tasks; averaged across all five tasks, this corresponds to 76.8\% for \BW{} versus 34.8\% for \BP{}. Within VR data, the wrench action space is significantly \emph{worse} on Fuse Clipping and shows no advantage on any other task, with 28.8\% average success for \VW{} compared to 40\% for \VP{}.

The cross-condition results isolate the contributions of the data collection interface and the action representation. Wrench actions trained on \VR (\VW) perform substantially worse than \BW on all tasks with 28.8\% average success, showing that using the wrench action space alone is insufficient if the training data does not contain intentionally commanded forces. Position policies trained on bilateral data (\BP) similarly underperform \BW, confirming that the bilateral interface alone does not recover the benefit without the matching action representation. Together, these results support our central claim: the alignment between data collection interface and action space is essential, and both components together lead to a significantly better policy performance.

\textbf{Contact Force Analysis.} Beyond task success, we evaluate whether direct wrench control produces more consistent physical interaction, i.e., less variance in the applied forces. For each rollout, we compute the 99th percentile of the measured end-effector force magnitude, which characterizes sustained near-peak contact loading while being less sensitive to isolated sensor spikes than the maximum. Figure~\ref{fig:contactForce} compares the distribution of this quantity across policies. Across the successful task-policy combinations, \BW achieves high success rates with lower median near-peak forces and/or a narrower distribution than the position-based alternatives. This suggests that bilateral wrench demonstrations allow the policy to execute contact interactions more consistently, rather than relying on large positional errors to generate incidental forces. Fan Insertion is an exception: although a comparison method exhibits a lower median force, its low success rate makes that lower loading uninformative because many trials do not complete the insertion.


\subsection{Inference Frequency Analysis}
\label{sec:ablations}
Force control is widely assumed to require high update rates, but this claim conflates two distinct loops. The inner torque control loop runs at a high frequency, whereas the outer policy loop, which needs only to track the slower timescale of task-relevant force variations. The mechanical impedance of the system, specifically the inertia of the end-effector and the damping in the controller, naturally attenuates high-frequency disturbances and prevents oscillation. We observe stable wrench control at \SI{50}{\hertz} policy inference with no oscillatory behavior. We further evaluate policy performance at reduced inference frequencies to determine how much this rate actually matters.

We evaluate this on the Industrial Connector task by reducing the policy inference frequency and executing the intervening control steps open loop (see Figure~\ref{fig:frequency}). At the nominal \SI{50}{\hertz} rate, \BW achieves an 80\% success rate. Performance decreases substantially at lower rates: 36\% at \SI{30}{\hertz}, 20\% at \SI{15}{\hertz}, 20\% at \SI{5}{\hertz}, and 12\% at \SI{1}{\hertz}. Each reduced-frequency condition is evaluated over 25 rollouts; the nominal-rate result is computed from 50 rollouts. These results show that stable wrench execution is possible with a policy operating well below the inner control-loop frequency, but that task performance remains sensitive to the temporal resolution of policy updates for this precise insertion task.

\subsection{Robustness to Box-Height Variation}
\label{sec:ablation_box_height}
We additionally evaluate the robustness of the trained Pen Writing policies by raising the box supporting the paper by \SI{2.5}{\centi\meter} (see Figure~\ref{fig:tasks}). The results are displayed in Table~\ref{tab:pen-writing-ablation}.
Our experiments show that even though the success rates of all policies diminish, the policies trained on the wrench action space tend to be more robust against height variations in the contact plane. We hypothesize this is due to the learned deliberate contact force, which, compared to an impedance controller, does not directly depend on the vertical spatial position.

\begin{table}[h]
  \centering
  \caption{Success rates (\%) for Pen Writing at the nominal box height and with the box raised by 2.5\,cm.}
  \label{tab:pen-writing-ablation}
  \begin{tabular}{lcccc}
  \toprule
  \textbf{Task} & \BW & \BP & \VW & \VP \\
  \midrule
  Pen Writing, Nominal Height             & 84 \% & 46\% & \textbf{86\%} & 78\% \\
  Pen Writing, Box Raised   & \textbf{76\%} & 8\% & 56\% & 24\% \\
  
  \bottomrule
  \end{tabular}
\end{table}

\section{Conclusions}
In this work, we demonstrate that direct wrench prediction is a viable action representation for contact-rich robotic manipulation. Across the tasks evaluated, the wrench-action policy trained on bilaterally collected data matches or outperforms the position-based baseline, supporting our hypothesis that wrenches are a natural and expressive action space for tasks where interaction forces are task-critical. By predicting desired contact wrenches directly, the policy can reason explicitly about what forces are needed rather than relying on emergent force behavior from a position controller.

Several natural next steps follow from this study. The most direct is scaling to a multi-task model trained jointly across tasks: the dataset of over 1{,}000 wrench-action demonstrations released alongside this work serves as a first step in that direction. Given enough wrench-action data, a multi-task model with a wrench action space should generalize across contact-rich tasks. Simulation offers one route to generating additional training data at scale, since wrench actions can be recorded from scripted or RL-trained experts in physics simulators without physical teleoperation. A separate question is whether existing datasets with force measurements but positional action spaces~\cite{manipforce,he2025foar,liu2025forcemimic,yu2026forcevla} can bootstrap a wrench model. Our results suggest caution, as post-hoc wrenches differ systematically from deliberately commanded ones.

\section*{Acknowledgment}
OpenAI’s ChatGPT and Anthropic’s Claude Code were used to assist with debugging and to generate limited amounts of boilerplate and data-processing code. All AI-generated code was reviewed, adapted, and validated by the authors. The research methodology, algorithm design, analysis, and interpretation of results were carried out by the authors. ChatGPT was also used to create illustrations shown in Figure \ref{fig:overview}.





\bibliographystyle{IEEEtran}
\bibliography{sources.bib}

\end{document}